\documentclass{article}

\usepackage{iclr2027_conference,times}

\usepackage{amsmath,amsfonts,bm}

\def\eqref#1{equation~\ref{#1}}
\def\1{\bm{1}}

\DeclareMathAlphabet{\mathsfit}{\encodingdefault}{\sfdefault}{m}{sl}
\SetMathAlphabet{\mathsfit}{bold}{\encodingdefault}{\sfdefault}{bx}{n}

\usepackage{hyperref}
\usepackage{url}
\usepackage{graphicx}
\usepackage{wrapfig}
\usepackage{booktabs}
\usepackage{amsmath}
\usepackage{amssymb}
\usepackage{multirow}
\usepackage{algorithm}
\usepackage{algpseudocode}
\usepackage{float}
\usepackage{marvosym}

\title{MUSE: Dependency-Aware Adaptation of a Frozen Vision Backbone for Multivariate Time Series Forecasting}

\author{Xinying Cai, Junkai Lu, Yuhan Zhu, Xiaoyun Yu, 
Xiangfei Qiu, Jilin Hu \\
East China Normal University
}

\iclrfinalcopy

\begin{document}

\maketitle

% arxiv
\lhead{Preprint}

% ==================================================
% Abstract
% ==================================================

\begin{abstract}
Multivariate time-series forecasting is essential to many real-world applications. Recent large vision models (LVMs) offer a promising paradigm by transferring cross-domain visual priors to time-series forecasting. However, existing LVM-based methods face two key challenges: balancing independent visual representation spaces with cross-variable dependency modeling, and adapting vision backbones pretrained on natural images to the distinct temporal semantics of time-series images. To address these challenges, we propose MUSE, a dependency-aware adaptation framework built on a fully frozen pretrained MAE. First, the Variable Context Refinement Module (VCR) aggregates shared temporal information within each variable and models cross-variable contextual dependencies while preserving independent visual spaces. Second, the Temporal--Periodic Refinement Module (TPR) performs lightweight refinement at different encoder depths and explicitly models across-period temporal dependencies and within-period periodic dependencies. The two modules independently produce forecasts, which are fused through a learnable prediction-level gate. Experiments on 10 real-world datasets demonstrate that MUSE achieves state-of-the-art performance.
\end{abstract}

% ==========================================
% Main Sections
% ==========================================

\section{Introduction}

Multivariate time-series forecasting aims to predict the future evolution of multiple correlated variables from historical observations, and plays a vital role in a wide range of real-world applications, including energy systems~\citep{hong2014gefcom,lago2021electricity}, traffic forecasting~\citep{li2018dcrnn,yu2018stgcn}, weather forecasting~\citep{bi2023pangu,lam2023graphcast}, and environmental monitoring~\citep{zheng2015airquality,wu2026airdde}. In recent years, numerous deep learning methods have been proposed and have substantially advanced time-series forecasting performance~\citep{qiu2026survey,salinas2020deepar,zhou2021informer,liu2022pyraformer,liu2022scinet,zhang2023crossformer}.

Furthermore, large-scale pretrained models provide time-series forecasting with prior knowledge beyond the numerical modality. Conventional methods primarily learn temporal patterns from limited numerical observations, whereas large-scale pretrained models, such as LLMs and LVMs, acquire rich semantic, structural, and contextual priors from large-scale cross-domain data, complementing information that is difficult to capture from numerical representations alone and thereby enhancing the representation and generalization capabilities of forecasting models~\citep{liu2025timecma,sun2026markovian,li2026tess,zhang2026timesaf,liang2024foundation,sun2024test,liu2024unitime}.

In particular, LVM-based forecasting methods have demonstrated remarkable cross-domain transferability. VisionTS~\citep{chen2025visionts} reformulates time-series forecasting as an image reconstruction task and directly leverages an MAE pretrained on natural images from ImageNet~\citep{deng2009imagenet,he2022mae}, achieving competitive zero-shot forecasting performance without further adaptation to the numerical time-series domain. Subsequent works, such as DMMV~\citep{shen2025dmmv} and VisionTS++~\citep{shen2025visiontspp}, further push forecasting performance to the state of the art. However, these methods still face two key challenges: how to balance independent visual representation spaces for individual variables with cross-variable dependency modeling, and how to better adapt visual representations to the distinctive structures induced by time-series imaging.

\begin{figure}[t]
    \centering
    \includegraphics[width=\linewidth]{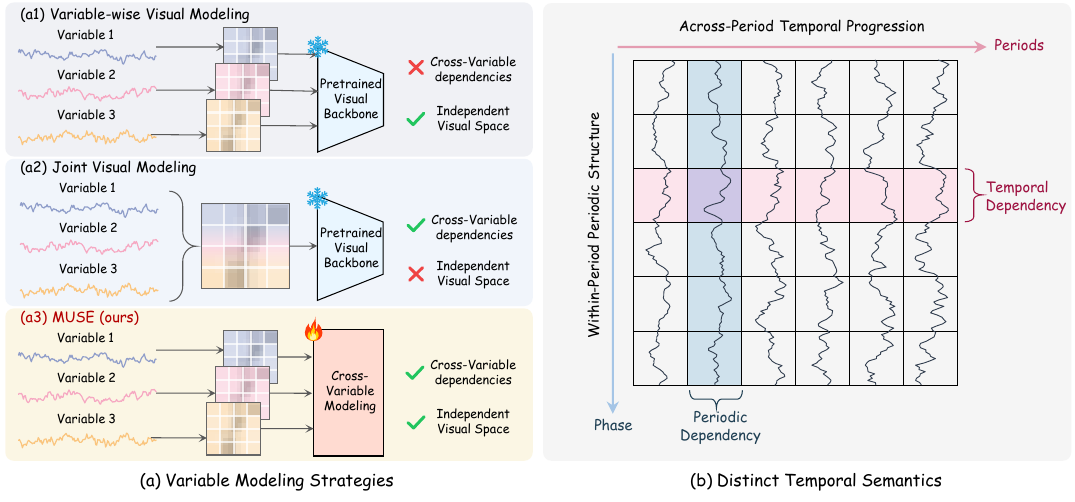}
    \caption{Motivation of MUSE. (a) Variable Modeling Strategies in the visual space. (b) Distinct temporal and periodic semantics of time-series images.}
    \label{fig:variable_modeling}
\end{figure}

First, there is an inherent trade-off between variable representation capacity and cross-variable dependency modeling. According to how multivariate time series are organized in the visual space, existing LVM-based forecasting methods can be broadly divided into two categories. 1) \textit{Variable-wise Visual Modeling} (Figure~\ref{fig:variable_modeling}(a1)): each variable is independently transformed into an image and processed by a shared visual backbone, as adopted by VisionTS~\citep{chen2025visionts}, ViTime~\citep{yang2025vitime}, and DMMV~\citep{shen2025dmmv}. This strategy allows each variable to fully exploit an independent visual representation space, but the isolated processing makes cross-variable dependencies difficult to capture explicitly. 2) \textit{Joint Visual Modeling} (Figure~\ref{fig:variable_modeling}(a2)): multiple variables are jointly organized into a single image, enabling their relationships to be modeled through shared visual representations, as in Time-VLM~\citep{zhong2025time} and VisionTS++~\citep{shen2025visiontspp}. However, jointly accommodating multiple variables within a fixed image space requires them to share limited visual representation capacity. Consequently, these two paradigms struggle to simultaneously achieve sufficient variable representation and effective cross-variable dependency modeling.

Second, a substantial structural semantic gap exists between natural images and time-series images. The two axes of natural images primarily encode spatial positions, whereas time-series images carry distinct temporal semantics: one captures temporal progression across periods, while the other characterizes periodic structure within each period, as illustrated in Figure~\ref{fig:variable_modeling}(b). An MAE pretrained on natural images is primarily optimized for modeling two-dimensional spatial relations and, when directly transferred to time-series images, lacks additional adaptation to their distinctive temporal--periodic structure, limiting its ability to fully capture temporal and periodic dependencies. Therefore, how to bridge this structural gap while keeping the visual backbone frozen and better adapt pretrained visual representations to time-series imaging constitutes the second key challenge.

Based on the above analysis, we propose \textbf{MUSE}, which performs \textit{Dependency-Aware Adaptation of a Frozen Vision Backbone for Multivariate Time Series Forecasting}. With the pretrained MAE fully frozen, MUSE adapts visual representations from two perspectives: \textit{variable-context dependencies and temporal--periodic dependencies}. Specifically, we first introduce the \textit{Variable Context Refinement Module (VCR)}, which aggregates shared temporal information within each variable into variable-level context and further models contextual dependencies across variables, thereby incorporating cross-variable information while preserving independent visual spaces. We then introduce the \textit{Temporal--Periodic Refinement Module (TPR)}, which performs lightweight adaptation at different depths of the frozen MAE Encoder through low-dimensional bottlenecks and models across-period temporal dependencies and within-period periodic dependencies, enabling pretrained visual representations to better accommodate the distinctive temporal--periodic structure of time-series images. Finally, the two modules independently produce forecasts, which are fused through a learnable prediction-level gate. Our main contributions are summarized as follows:

\begin{itemize}
    \item We propose \textbf{MUSE}, a dependency-aware adaptation framework that transfers pretrained visual priors to multivariate time-series forecasting with a fully frozen visual backbone.
    \item We introduce VCR to model variable-context dependencies while preserving independent visual representation spaces.
    \item We introduce TPR, a lightweight bottleneck-based adaptation module that captures temporal and periodic dependencies to bridge the structural semantic gap between natural and time-series images.
    \item We conduct extensive experiments on 10 multivariate time-series datasets. The results show that MUSE consistently outperforms representative state-of-the-art baselines. Additionally, the code is available at \url{https://github.com/decisionintelligence/MUSE}.
\end{itemize}

% Temporary numbering before Sections 1 is added
% \setcounter{section}{1}

\section{Related Works}

\subsection{Multivariate Time Series Forecasting}

Multivariate time series forecasting has been extensively studied. Early approaches mainly relied on statistical modeling~\citep{sims1980macroeconomics}, followed by machine learning methods for capturing nonlinear temporal patterns~\citep{sapankevych2009timeseries,ahmed2010empirical}. In recent years, deep learning has become the dominant paradigm. Representative methods include LSTNet~\citep{lai2018modeling}, Autoformer~\citep{wu2021autoformer}, FEDformer~\citep{zhou2022fedformer}, DLinear~\citep{zeng2023dlinear}, PatchTST~\citep{nie2023patchtst}, TimesNet~\citep{wu2023timesnet}, iTransformer~\citep{liu2024itransformer}, Pathformer~\citep{chen2024pathformer}, TimeMixer~\citep{wang2024timemixer}, and DUET~\citep{qiu2025duet}. These methods have continuously improved forecasting performance, but are typically trained end-to-end from scratch on numerical time series, with limited use of the rich prior knowledge provided by large-scale cross-domain pretrained models.

\subsection{LVM-Based Time Series Forecasting}

With the development of visual models such as ResNet~\citep{he2016resnet}, VGG~\citep{simonyan2015vgg}, Inception~\citep{szegedy2015inception}, ViT~\citep{dosovitskiy2021vit}, Swin Transformer~\citep{liu2021swin}, and MAE~\citep{he2022mae}, transferring visual representations to time-series forecasting has attracted increasing attention. One line of work adopts channel-independent modeling. ViTime~\citep{yang2025vitime} explores a vision-based foundation model for time-series forecasting. VisionTS~\citep{chen2025visionts} reformulates forecasting as masked image reconstruction and leverages a pretrained MAE for prediction. DMMV~\citep{shen2025dmmv} builds upon VisionTS by introducing a numerical branch for complementary forecasting. Aurora~\citep{wu2026aurora} develops a cross-domain multimodal time-series foundation model for zero-shot forecasting. Another line of work adopts channel-dependent modeling. Time-VLM~\citep{zhong2025time} and VisionTS++~\citep{shen2025visiontspp} both encode multiple variables into a single image to capture inter-variable relationships.

Overall, channel-independent methods preserve a complete visual representation space for each variable but lack explicit cross-variable interaction, while channel-dependent methods capture variable relationships at the cost of sharing a limited image space across variables. MUSE addresses both limitations by constructing an individual image for each variable while explicitly modeling inter-variable dependencies, and further refines temporal and periodic structures through lightweight adaptation over a fully frozen visual backbone.
\section{Methodology}

\begin{figure}[t]
    \centering
    \includegraphics[width=\linewidth]{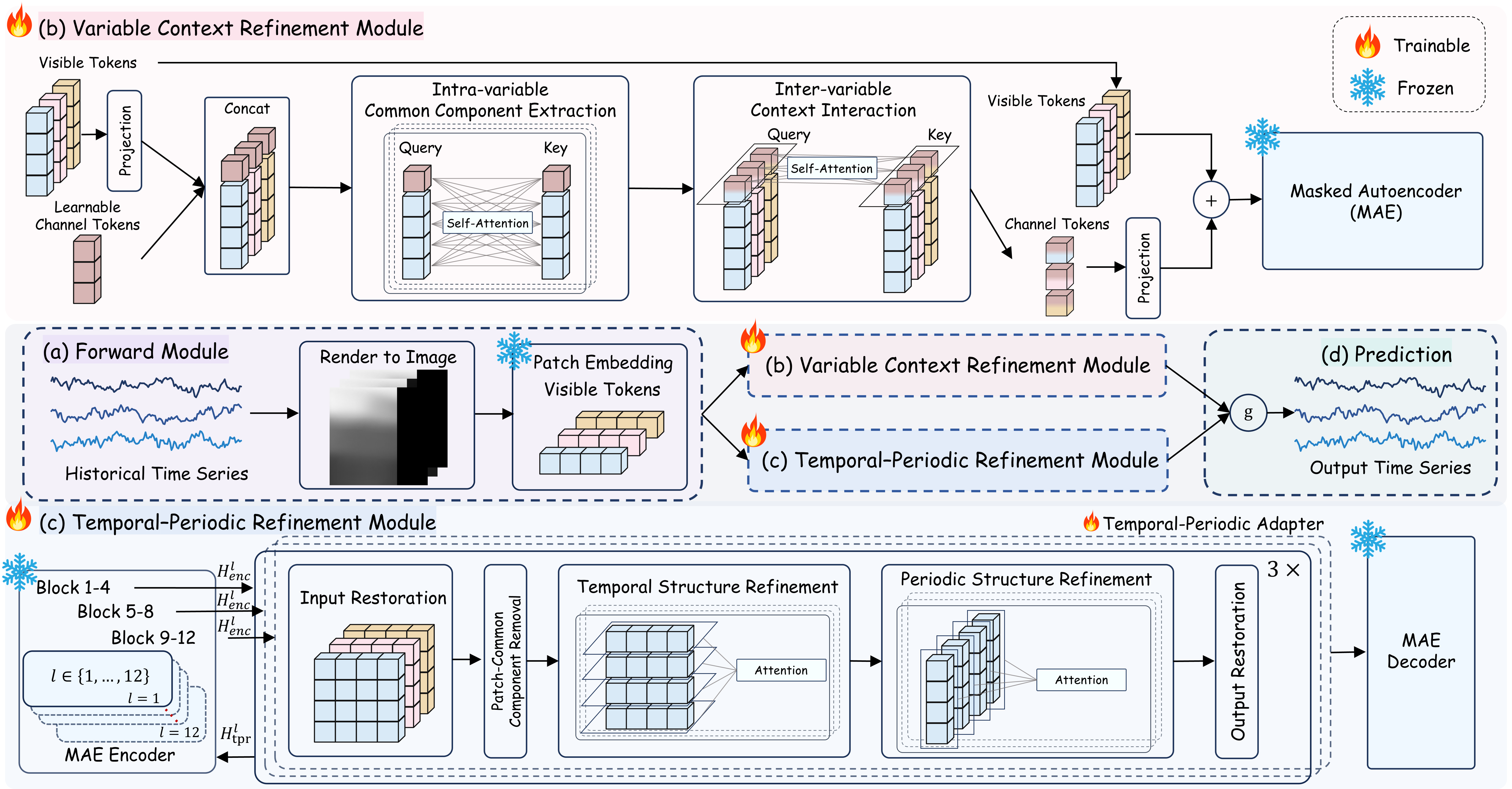}
    \caption{Overall architecture of MUSE. The historical multivariate time series is first processed by the Forward Module to obtain visible historical tokens, which are then fed in parallel into the Variable Context Refinement Module (VCR) and the Temporal--Periodic Refinement Module (TPR). VCR models variable context, while TPR refines temporal--periodic structures. The predictions of the two modules are finally fused through a learnable prediction-level gate.}
    \label{fig:muse_overview}
\end{figure}

\paragraph{Problem Statement.}

Given a multivariate time series with $V$ variables, the historical observations within a look-back window of length $L$ are denoted as $\mathbf{X}=[\mathbf{x}_{t-L+1},\ldots,\mathbf{x}_{t}]\in\mathbb{R}^{L\times V}$, where $\mathbf{x}_{t}\in\mathbb{R}^{V}$ denotes the observations of all variables at time step $t$. Multivariate time-series forecasting learns a mapping $f:\mathbb{R}^{L\times V}\rightarrow\mathbb{R}^{H\times V}$ to predict the next $H$ steps, yielding the prediction, $\widehat{\mathbf{Y}}=f(\mathbf{X})\in\mathbb{R}^{H\times V}$, with ground truth $\mathbf{Y}=[\mathbf{x}_{t+1},\ldots,\mathbf{x}_{t+H}]\in\mathbb{R}^{H\times V}$.

\subsection{Structure Overview}

As illustrated in Figure~\ref{fig:muse_overview}, MUSE is built on a fully frozen pretrained MAE~\citep{he2022mae}. First, the historical multivariate time series is processed by the Forward Module to obtain visible historical tokens, which are then fed into two complementary structure-aware modules: the \textbf{Variable Context Refinement Module (VCR)} and the \textbf{Temporal--Periodic Refinement Module (TPR)}.

VCR aggregates intra-variable information and models cross-variable dependencies to enrich the original visual representations with variable context, while TPR refines temporal and periodic structures at different depths of the frozen MAE Encoder. The two modules independently produce forecasts, which are fused through a learnable prediction-level gate. The details of each module will be introduced in the following sections.

\subsection{Forward Module}

As in existing LVM-based time-series forecasting methods such as VisionTS and DMMV~\citep{chen2025visionts, shen2025dmmv}, MUSE transforms each variable into a 2D visual representation through the Forward Module. A detailed description is provided in Appendix~\ref{app:forward_module}.

Let $G$ denote the number of patches along each spatial axis and $K_{\mathrm{in}}$ the number of historical patch columns, giving $P=GK_{\mathrm{in}}$ visible historical patches per variable. After the frozen Patch Embedding, the visible historical tokens are denoted as $\mathbf{T}\in\mathbb{R}^{V\times P\times E}$, where $E$ is the MAE Encoder embedding dimension. $\mathbf{T}$ serves as the common input to VCR and TPR.

\subsection{Variable Context Refinement Module}
\label{sec:vcr}

Historical patch tokens within each variable often contain shared temporal information, while dependencies may also exist across variables. However, period-based imaging processes variables independently, without explicitly aggregating such intra-variable information or modeling cross-variable dependencies, resulting in visual representations that lack variable-level context. Inspired by TimeXer~\citep{wang2024timexer}, we introduce the Variable Context Refinement Module (VCR), which aggregates shared temporal information within each variable, models cross-variable dependencies, and uses the resulting variable context to refine the original patch representations.

Given the historical representation $\mathbf{T}\in\mathbb{R}^{V\times P\times E}$ produced by the frozen Patch Embedding, VCR first projects it into a low-dimensional feature space of dimension $D$:
\begin{equation}
    \mathbf{M}
    =
    \operatorname{LayerNorm}
    \left(
        \mathbf{W}_{d}
        \operatorname{LayerNorm}(\mathbf{T})
    \right),
\end{equation}
where $\mathbf{W}_{d}:\mathbb{R}^{E}\rightarrow\mathbb{R}^{D}$ denotes a trainable linear projection. The resulting patch memory $\mathbf{M}\in\mathbb{R}^{V\times P\times D}$ is used for subsequent variable-level information modeling, while the original visual representation $\mathbf{T}$ is retained for the final residual refinement.

\paragraph{Intra-variable Common Component Extraction.}
To capture temporal patterns shared across historical patch tokens within each variable, we introduce a shared learnable channel token $\mathbf{q}\in\mathbb{R}^{D}$ and replicate it across all variables, yielding $\mathbf{Q}=[\mathbf{q}_{1},\ldots,\mathbf{q}_{V}]\in\mathbb{R}^{V\times D}$. For the $v$-th variable, its channel token $\mathbf{q}_{v}$ is concatenated with the corresponding patch memory $\mathbf{M}_{v}\in\mathbb{R}^{P\times D}$ to form $\mathbf{S}_{v}=[\mathbf{q}_{v};\mathbf{M}_{v}]\in\mathbb{R}^{(1+P)\times D}$.

Multi-head self-attention is then performed independently within each variable:
\begin{equation}
    [\widetilde{\mathbf{q}}_{v};\widetilde{\mathbf{M}}_{v}]
    =
    \operatorname{LayerNorm}
    \left(
        \mathbf{S}_{v}
        +
        \operatorname{MultiHeadSelfAttention}_{\mathrm{intra}}
        \left(
            \operatorname{LayerNorm}(\mathbf{S}_{v})
        \right)
    \right).
\end{equation}
Through interaction with all historical patch tokens of the corresponding variable, the updated channel token $\widetilde{\mathbf{q}}_{v}\in\mathbb{R}^{D}$ aggregates shared temporal patterns into a compact variable-level representation. The same intra-variable self-attention parameters are shared across all variables.

\paragraph{Inter-variable Context Interaction.}
After obtaining a compact representation for each variable, we further model dependencies across variables. The updated channel tokens are organized as a variable-level sequence $\widetilde{\mathbf{Q}}=[\widetilde{\mathbf{q}}_{1},\ldots,\widetilde{\mathbf{q}}_{V}]\in\mathbb{R}^{V\times D}$, and multi-head self-attention is performed along the variable dimension:
\begin{equation}
    \mathbf{Q}'
    =
    \operatorname{LayerNorm}
    \left(
        \widetilde{\mathbf{Q}}
        +
        \operatorname{MultiHeadSelfAttention}_{\mathrm{inter}}
        \left(
            \operatorname{LayerNorm}(\widetilde{\mathbf{Q}})
        \right)
    \right).
\end{equation}
This yields $\mathbf{Q}'=[\mathbf{q}'_{1},\ldots,\mathbf{q}'_{V}]\in\mathbb{R}^{V\times D}$, where each $\mathbf{q}'_{v}$ incorporates contextual information from other variables.

The resulting channel token $\mathbf{q}'_{v}\in\mathbb{R}^{D}$ is transformed by a trainable write-back projection and mapped back to the MAE embedding space, producing a variable-specific correction $\mathbf{C}_{v}\in\mathbb{R}^{E}$:
\begin{equation}
    \mathbf{C}_{v}
    =
    \operatorname{LinearProjection}_{D\rightarrow E}
    \left(
        \operatorname{WriteBackProjection}
        (\mathbf{q}'_{v})
    \right).
\end{equation}
Each $\mathbf{C}_{v}$ is broadcast along the patch dimension within the corresponding variable:
\begin{equation}
    \mathbf{T}^{\mathrm{vcr}}_{v,p}
    =
    \mathbf{T}_{v,p}
    +
    \mathbf{C}_{v},
    \qquad
    p=1,\ldots,P.
\end{equation}
This yields $\mathbf{T}^{\mathrm{vcr}}\in\mathbb{R}^{V\times P\times E}$, where each variable receives its own shared correction across historical patches. In this way, VCR incorporates intra-variable shared patterns and inter-variable context while preserving the original patch-level local structure.

After prepending the frozen CLS token, $\mathbf{T}^{\mathrm{vcr}}$ is processed by the frozen MAE Encoder and Decoder, yielding the VCR prediction $\widehat{\mathbf{Y}}_{\mathrm{vcr}}\in\mathbb{R}^{H\times V}$.

\subsection{Temporal--Periodic Refinement Module}
\label{sec:tpr}

Period-based imaging converts one-dimensional time series into two-dimensional representations, preserving temporal information along both temporal and periodic axes. However, pretrained MAE self-attention is designed for natural images and does not explicitly distinguish these two axes with different temporal semantics. We therefore introduce the Temporal--Periodic Refinement Module (TPR) to model temporal dependencies across historical periods and periodic dependencies within each period. Inspired by parameter-efficient adaptation methods in computer vision such as AdaptFormer~\citep{chen2022adaptformer}, TPR performs lightweight refinement through a bottleneck architecture in a low-dimensional space.

Unlike VCR, TPR directly operates on the original historical representation $\mathbf{T}$ with a channel-independent design. The frozen CLS token is prepended before the MAE Encoder. For depth-wise adaptation, the 12 frozen Encoder Blocks are divided into three stages: Blocks 1--4, 5--8, and 9--12. Each stage uses an independent Temporal--Periodic Adapter shared across its four Blocks.

\paragraph{Input Restoration.}

After the $l$-th Frozen Encoder Block, $l\in\{1,\ldots,12\}$, let
$\mathbf{H}_{\mathrm{enc}}^{l}\in\mathbb{R}^{V\times(1+P)\times E}$
denote its output. The CLS token is retained unchanged, while the historical patch tokens in $\mathbf{H}_{\mathrm{enc}}^{l}$ are restored to their original 2D arrangement, yielding
$\mathbf{X}^{l}\in\mathbb{R}^{V\times G\times K_{\mathrm{in}}\times E}$. Each $(r,c)$ indexes a temporal--periodic patch location for subsequent modeling along both axes.

\paragraph{Patch-Common Component Removal.}
To emphasize temporal--periodic variations, TPR removes the representation component shared across historical patches. For the $v$-th variable, the mean representation is computed as
\begin{equation}
    \overline{\mathbf{X}}_{v}^{\,l}
    =
    \frac{1}{G K_{\mathrm{in}}}
    \sum_{r=1}^{G}
    \sum_{c=1}^{K_{\mathrm{in}}}
    \mathbf{X}_{v,r,c}^{l},
\end{equation}
and subtracted from each patch:
\begin{equation}
    \mathbf{R}_{v,r,c}^{l}
    =
    \mathbf{X}_{v,r,c}^{l}
    -
    \overline{\mathbf{X}}_{v}^{\,l}.
\end{equation}
The centered representation $\mathbf{R}^{l}\in\mathbb{R}^{V\times G\times K_{\mathrm{in}}\times E}$ is used for subsequent structural modeling, while $\mathbf{X}^{l}$ is retained for residual write-back.

$\mathbf{R}^{l}$ is then projected into a low-dimensional space of dimension $D_{\mathrm{tp}}$:
\begin{equation}
    \mathbf{Z}^{l}
    =
    \mathbf{W}_{\mathrm{down}}
    \left(
        \operatorname{LayerNorm}
        \left(
            \mathbf{R}^{l}
        \right)
    \right),
\end{equation}
where $\mathbf{W}_{\mathrm{down}}:\mathbb{R}^{E}\rightarrow\mathbb{R}^{D_{\mathrm{tp}}}$ is the down-projection of the stage-specific Temporal--Periodic Adapter, yielding $\mathbf{Z}^{l}\in\mathbb{R}^{V\times G\times K_{\mathrm{in}}\times D_{\mathrm{tp}}}$.

\paragraph{Temporal Structure Refinement.}
Temporal Attention is applied along the historical time axis across periods. For a fixed variable $v$ and periodic position $r$, the sequence $[\mathbf{Z}^{l}_{v,r,1},\ldots,\mathbf{Z}^{l}_{v,r,K_{\mathrm{in}}}]$ represents the same periodic position across consecutive historical periods, capturing its temporal progression over time. Multi-head self-attention is performed within this sequence.

To encode relative temporal relationships, for query position $c_i$ and key position $c_j$, we define the directional offset $d_t=c_i-c_j$. The attention logit of the $h$-th head is
\begin{equation}
    \ell_{ij}^{t,(h)}
    =
    \frac{
        \left\langle
            \mathbf{q}_{i}^{t,(h)},
            \mathbf{k}_{j}^{t,(h)}
        \right\rangle
    }{
        \sqrt{d_h}
    }
    +
    \mathbf{B}_{t}^{(h)}[d_t],
\end{equation}
where $\mathbf{q}_{i}^{t,(h)}$ and $\mathbf{k}_{j}^{t,(h)}$ denote the query and key vectors of the $h$-th Temporal Attention head, $d_h$ is the head dimension, and $\mathbf{B}_{t}^{(h)}$ is a learnable temporal relative-position bias. The offset $d_t$ encodes both relative distance and direction. The attention output is residually added to $\mathbf{Z}^{l}$, yielding $\mathbf{Z}_{t}^{l}$.

\paragraph{Periodic Structure Refinement.}
After Temporal Attention, Periodic Attention is applied along the phase axis within each historical period. For a fixed variable $v$ and historical column $c$, the sequence $[\mathbf{Z}^{l}_{t,v,1,c},\ldots,\mathbf{Z}^{l}_{t,v,G,c}]$ represents different periodic positions within the same historical period segment, capturing their within-period dependencies. Multi-head self-attention is performed within this sequence.

To preserve the periodic wrap-around continuity, for periodic positions $r_i$ and $r_j$, we define the directional circular offset as $d_p=(r_i-r_j)\bmod G$. The attention logit of the $h$-th head is
\begin{equation}
    \ell_{ij}^{p,(h)}
    =
    \frac{
        \left\langle
            \mathbf{q}_{i}^{p,(h)},
            \mathbf{k}_{j}^{p,(h)}
        \right\rangle
    }{
        \sqrt{d_h}
    }
    +
    \mathbf{B}_{p}^{(h)}[d_p],
\end{equation}
where $\mathbf{q}_{i}^{p,(h)}$ and $\mathbf{k}_{j}^{p,(h)}$ denote the query and key vectors of the $h$-th Periodic Attention head, and $\mathbf{B}_{p}^{(h)}$ is a learnable periodic relative-position bias. The circular offset preserves the adjacency between the beginning and end of a period while encoding directional relationships among periodic positions. The attention output is residually added to $\mathbf{Z}_{t}^{l}$, yielding $\mathbf{Z}_{\mathrm{tp}}^{l}$.

\paragraph{Output Restoration.}
For the $l$-th Encoder Block, the refined low-dimensional representation is projected back to the MAE embedding space:
\begin{equation}
    \Delta\mathbf{X}_{\mathrm{raw}}^{l}
    =
    \mathbf{W}_{\mathrm{up}}
    \left(
        \operatorname{LayerNorm}
        \left(
            \mathbf{Z}_{\mathrm{tp}}^{l}
        \right)
    \right),
\end{equation}
where $\mathbf{W}_{\mathrm{up}}:\mathbb{R}^{D_{\mathrm{tp}}}\rightarrow\mathbb{R}^{E}$ is the up-projection of the stage-specific Temporal--Periodic Adapter.

The refinement strength is controlled by a learnable parameter $s_l$, and the resulting correction is injected through a residual connection:
\begin{equation}
    \mathbf{X}_{\mathrm{out}}^{l}
    =
    \mathbf{X}^{l}
    +
    \operatorname{Sigmoid}(s_l)\,
    \Delta\mathbf{X}_{\mathrm{raw}}^{l}.
\end{equation}

The refined representation $\mathbf{X}_{\mathrm{out}}^{l}\in\mathbb{R}^{V\times G\times K_{\mathrm{in}}\times E}$ is flattened to the original patch order and combined with the retained CLS token, forming $\mathbf{H}_{\mathrm{tpr}}^{l}\in\mathbb{R}^{V\times(1+P)\times E}$ for the next Frozen Encoder Block when $l<12$.

After the final Encoder Block and TPR refinement, the representation passes through the Frozen Encoder final LayerNorm and the shared Frozen MAE Decoder to reconstruct the masked future region in the image space. The reconstruction is converted back to the time-series domain, yielding the TPR prediction $\widehat{\mathbf{Y}}_{\mathrm{tpr}}\in\mathbb{R}^{H\times V}$.

\subsection{Prediction}

VCR and TPR produce predictions $\widehat{\mathbf{Y}}_{\mathrm{vcr}}$ and $\widehat{\mathbf{Y}}_{\mathrm{tpr}}$, respectively. We use a learnable prediction-level gate $g=\operatorname{Sigmoid}(a)$, where $a$ is a global learnable parameter. The final prediction is
\begin{equation}
    \widehat{\mathbf{Y}}
    =
    g\,\widehat{\mathbf{Y}}_{\mathrm{vcr}}
    +
    (1-g)\widehat{\mathbf{Y}}_{\mathrm{tpr}}.
\end{equation}

We use MSE losses $\mathcal{L}_{\mathrm{vcr}}$, $\mathcal{L}_{\mathrm{tpr}}$, and $\mathcal{L}_{\mathrm{gate}}$ to supervise VCR, TPR, and the gate, respectively, where the gate is optimized using detached branch predictions. The overall objective is
\begin{equation}
    \mathcal{L}
    =
    \mathcal{L}_{\mathrm{vcr}}
    +
    \mathcal{L}_{\mathrm{tpr}}
    +
    \mathcal{L}_{\mathrm{gate}}.
\end{equation}
\section{Experiments}

\subsection{Experimental Settings}
\label{sec:experimental_settings}

\paragraph{Datasets.}
We evaluate MUSE on 10 multivariate time-series datasets from the Time Series Forecasting Benchmark (TFB)~\citep{qiu2024tfb}, including ETTh1, ETTh2, ETTm1, ETTm2, Weather, Electricity, AQWan, CzeLan, Wind, and ZafNoo, covering electricity, weather, air quality, wind energy, and natural environments. We follow the standard train/validation/test splits, using $60\%/20\%/20\%$ for the ETT datasets and AQWan, and $70\%/10\%/20\%$ for the remaining datasets~\citep{qiu2024tfb}. Further dataset details are provided in Appendix~\ref{app:datasets}.

\paragraph{Baselines.}
We compare MUSE with eight representative state-of-the-art forecasting models, including the multimodal model DMMV-A \citep{shen2025dmmv}, the LVM-based model VisionTS \citep{chen2025visionts}, the LLM-based models Time-LLM \citep{jin2024timellm} and GPT4TS \citep{zhou2023onefitsall}, and numerical time-series models DLinear \citep{zeng2023dlinear}, PatchTST \citep{nie2023patchtst}, TimesNet \citep{wu2023timesnet}, and FEDformer \citep{zhou2022fedformer}.

\paragraph{Implementation Details.}
We use Mean Squared Error (MSE) and Mean Absolute Error (MAE) as the evaluation metrics and uniformly set the forecasting horizon to $H\in\{96,192,336,720\}$ for all datasets. All experiments of MUSE are conducted under the unified evaluation framework of the Time Series Forecasting Benchmark (TFB) \citep{qiu2024tfb}. All experiments of MUSE are implemented using PyTorch in Python 3.11.15 and executed on NVIDIA H20 GPUs. Further implementation details are provided in Appendix~\ref{app:implementation_details}.

\subsection{Main Results}
\begin{table*}[t]
    \centering
    \caption{Average forecasting results on 10 multivariate time-series datasets over four horizons $H\in\{96,192,336,720\}$. \textcolor{red}{\textbf{Red}}: the best, \textcolor{blue}{\underline{Blue}}: the 2nd best. Full results are available in Table~\ref{tab:main_results_full} in Appendix~\ref{app:full_results}. Lower is better.}
    \label{tab:main_results_avg}
    \small
    \setlength{\tabcolsep}{2.0pt}
    \renewcommand{\arraystretch}{1.08}
    \resizebox{\textwidth}{!}{%
    \begin{tabular}{l*{9}{cc}}
        \toprule
        \textbf{Models} & \multicolumn{2}{c}{MUSE (ours)} & \multicolumn{2}{c}{DMMV-A} & \multicolumn{2}{c}{VisionTS} & \multicolumn{2}{c}{Time-LLM} & \multicolumn{2}{c}{GPT4TS} & \multicolumn{2}{c}{DLinear} & \multicolumn{2}{c}{PatchTST} & \multicolumn{2}{c}{TimesNet} & \multicolumn{2}{c}{FEDformer} \\
        \textbf{Metrics} & MSE & MAE & MSE & MAE & MSE & MAE & MSE & MAE & MSE & MAE & MSE & MAE & MSE & MAE & MSE & MAE & MSE & MAE \\
        \midrule
        ETTh1 & \textcolor{red}{\textbf{0.384}} & \textcolor{red}{\textbf{0.405}} & 0.395 & \textcolor{blue}{\underline{0.414}} & \textcolor{blue}{\underline{0.388}} & \textcolor{blue}{\underline{0.414}} & 0.418 & 0.432 & 0.418 & 0.421 & 0.423 & 0.437 & 0.413 & 0.431 & 0.458 & 0.450 & 0.440 & 0.460 \\
        \midrule
        ETTh2 & \textcolor{red}{\textbf{0.325}} & \textcolor{red}{\textbf{0.372}} & 0.337 & 0.388 & 0.339 & 0.384 & 0.361 & 0.396 & 0.354 & 0.389 & 0.431 & 0.447 & \textcolor{blue}{\underline{0.330}} & \textcolor{blue}{\underline{0.379}} & 0.414 & 0.427 & 0.437 & 0.449 \\
        \midrule
        ETTm1 & \textcolor{red}{\textbf{0.330}} & \textcolor{red}{\textbf{0.362}} & 0.340 & 0.371 & \textcolor{blue}{\underline{0.338}} & \textcolor{blue}{\underline{0.367}} & 0.356 & 0.377 & 0.363 & 0.378 & 0.357 & 0.379 & 0.351 & 0.381 & 0.400 & 0.406 & 0.448 & 0.452 \\
        \midrule
        ETTm2 & \textcolor{red}{\textbf{0.245}} & \textcolor{red}{\textbf{0.309}} & 0.256 & 0.317 & 0.265 & 0.324 & 0.261 & 0.316 & \textcolor{blue}{\underline{0.254}} & \textcolor{blue}{\underline{0.311}} & 0.267 & 0.334 & 0.255 & 0.315 & 0.291 & 0.333 & 0.305 & 0.349 \\
        \midrule
        Electricity & \textcolor{red}{\textbf{0.157}} & \textcolor{blue}{\underline{0.251}} & \textcolor{blue}{\underline{0.158}} & \textcolor{red}{\textbf{0.248}} & 0.164 & 0.257 & 0.165 & 0.259 & 0.170 & 0.263 & 0.166 & 0.264 & 0.162 & 0.253 & 0.193 & 0.295 & 0.214 & 0.327 \\
        \midrule
        Wind & \textcolor{red}{\textbf{1.018}} & \textcolor{red}{\textbf{0.735}} & 1.140 & \textcolor{blue}{\underline{0.751}} & 1.121 & 0.760 & 1.173 & 0.780 & 1.220 & 0.768 & \textcolor{blue}{\underline{1.077}} & \textcolor{red}{\textbf{0.735}} & 1.120 & 0.765 & 1.251 & 0.781 & 1.254 & 0.815 \\
        \midrule
        Weather & \textcolor{red}{\textbf{0.216}} & 0.261 & \textcolor{blue}{\underline{0.217}} & \textcolor{blue}{\underline{0.256}} & 0.262 & 0.293 & 0.244 & 0.270 & 0.227 & \textcolor{red}{\textbf{0.255}} & 0.249 & 0.300 & 0.226 & 0.264 & 0.259 & 0.287 & 0.309 & 0.360 \\
        \midrule
        AQWan & 0.805 & \textcolor{red}{\textbf{0.494}} & \textcolor{blue}{\underline{0.804}} & 0.528 & 0.826 & 0.507 & \textcolor{red}{\textbf{0.800}} & 0.510 & 0.819 & \textcolor{red}{\textbf{0.494}} & 0.818 & 0.512 & 0.812 & \textcolor{blue}{\underline{0.499}} & 0.813 & 0.501 & 0.848 & 0.529 \\
        \midrule
        ZafNoo & \textcolor{red}{\textbf{0.494}} & \textcolor{red}{\textbf{0.432}} & 0.507 & \textcolor{blue}{\underline{0.450}} & 0.538 & 0.452 & 0.561 & 0.483 & 0.547 & 0.456 & \textcolor{blue}{\underline{0.496}} & 0.451 & 0.512 & 0.465 & 0.537 & 0.465 & 0.578 & 0.499 \\
        \midrule
        CzeLan & \textcolor{red}{\textbf{0.211}} & \textcolor{red}{\textbf{0.264}} & 0.226 & 0.294 & 0.241 & 0.290 & 0.229 & 0.285 & 0.248 & \textcolor{blue}{\underline{0.271}} & 0.285 & 0.343 & 0.227 & 0.290 & \textcolor{blue}{\underline{0.224}} & 0.285 & 0.303 & 0.364 \\
        \midrule
        1$^{\mathrm{st}}$ Count & \textcolor{red}{\textbf{9}} & \textcolor{red}{\textbf{8}} & 0 & 1 & 0 & 0 & \textcolor{blue}{\underline{1}} & 0 & 0 & \textcolor{blue}{\underline{2}} & 0 & 1 & 0 & 0 & 0 & 0 & 0 & 0 \\
        \bottomrule
    \end{tabular}%
    }
\end{table*}

Comprehensive forecasting results are summarized in Table~\ref{tab:main_results_avg}, with full results over all forecasting horizons provided in Appendix~\ref{app:full_results}. We make the following observations. 1) MUSE shows an overall advantage. Across the averaged results on 10 datasets, MUSE ranks first on 9 MSE and 8 MAE metrics. The remaining results also stay among the leading methods. 2) MUSE substantially improves the adaptation of pretrained visual models to time-series forecasting. Compared with existing visual-pretraining-based methods (e.g., VisionTS and DMMV-A), MUSE performs better on most datasets. Notably, \textbf{these gains are achieved with the MAE backbone fully frozen}. Using only lightweight structure-aware modules for temporal adaptation, MUSE effectively adapts frozen visual representations to multivariate forecasting. 3) Across 10 datasets spanning diverse domains and temporal characteristics, MUSE remains competitive against multimodal, LVM-based, LLM-based, and numerical forecasters. The consistent gains across these heterogeneous settings demonstrate stable performance and effective transfer of the cross-domain priors of the frozen visual backbone.

\subsection{Ablation Studies}

\begin{table*}[t]
    \centering
    \caption{Ablation studies for MUSE. TP denotes Temporal--Periodic. Lower is better; the best results are in bold.}
    \label{tab:ablation_main}
    \small
    \setlength{\tabcolsep}{2.2pt}
    \renewcommand{\arraystretch}{1.12}
    \begin{tabular*}{\textwidth}{@{\extracolsep{\fill}}lccccc@{}}
        \toprule
        \multirow{2}{*}{Dataset}
        & Full Model
        & w/o VCR
        & w/o TPR
        & w/o PCR
        & Replace TP Attention \\
        & MSE / MAE
        & MSE / MAE
        & MSE / MAE
        & MSE / MAE
        & MSE / MAE \\
        \midrule
        ETTh2
        & \textbf{0.373 / 0.416}
        & 0.380 / 0.422
        & 0.378 / 0.423
        & 0.374 / 0.418
        & 0.388 / 0.430 \\
        Weather
        & \textbf{0.299 / 0.326}
        & 0.309 / 0.330
        & 0.300 / 0.331
        & 0.300 / 0.328
        & 0.301 / 0.327 \\
        AQWan
        & \textbf{0.883 / 0.526}
        & 0.899 / 0.535
        & 0.893 / 0.535
        & 0.888 / 0.532
        & 0.887 / 0.529 \\
        ZafNoo
        & \textbf{0.564 / 0.468}
        & 0.571 / 0.471
        & 0.575 / 0.473
        & 0.565 / 0.469
        & 0.565 / 0.470 \\
        \midrule
        Average
        & \textbf{0.530 / 0.434}
        & 0.540 / 0.440
        & 0.537 / 0.441
        & 0.532 / 0.437
        & 0.535 / 0.439 \\
        \bottomrule
    \end{tabular*}
\end{table*}

To assess the contribution of each key design in MUSE, we conduct ablation studies on four representative datasets at $H=720$. We compare the full model with four variants: 1) \textit{w/o VCR}: removes the Variable Context Refinement Module; 2) \textit{w/o TPR}: removes the Temporal--Periodic Refinement Module; 3) \textit{w/o PCR}: removes Patch-Common Component Removal from TPR in Section~\ref{sec:tpr}; and 4) \textit{Replace TP Attention}: replaces the Temporal--Periodic (TP) Attention, comprising Temporal Attention in Temporal Structure Refinement and Periodic Attention in Periodic Structure Refinement of TPR (Section~\ref{sec:tpr}), with mean pooling along their corresponding axes.

As shown in Table~\ref{tab:ablation_main}, the full model performs best on all datasets and on average. Removing VCR or TPR increases the forecasting error, confirming the effectiveness of both modules. Removing PCR also degrades performance, demonstrating the benefit of eliminating components shared across historical patches. Replacing TP Attention with axis-wise mean pooling degrades performance, showing that adaptive attention captures temporal and periodic dependencies more effectively than uniform aggregation. Overall, MUSE benefits jointly from variable-context modeling, patch-common component removal, and adaptive temporal--periodic refinement, validating the overall design.

\subsection{Parameter Sensitivity}

\begin{figure}[t]
    \centering
    \includegraphics[width=\linewidth]{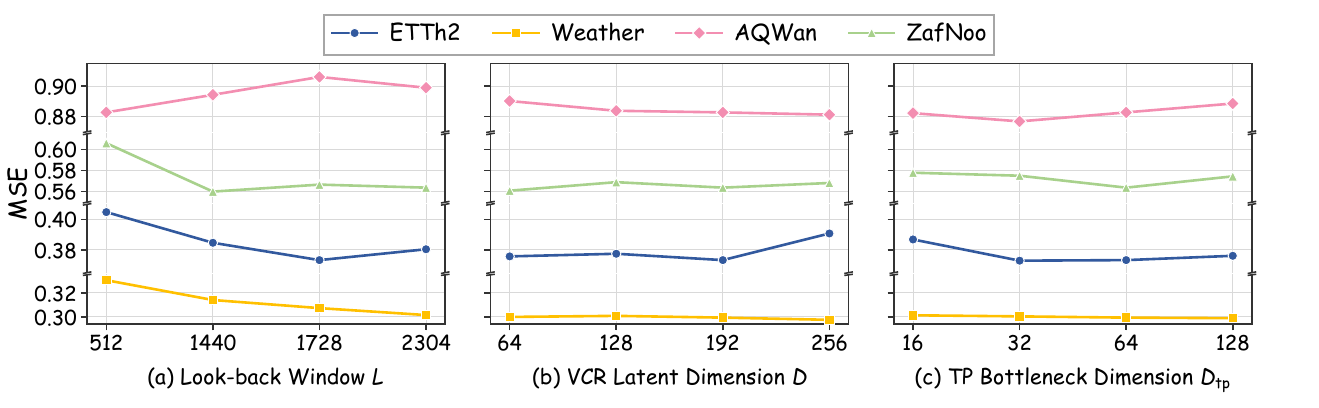}
    \caption{Parameter sensitivity studies of main hyper-parameters in MUSE. The broken vertical axes omit ranges containing no observations; lower is better.}
    \label{fig:parameter_sensitivity}
\end{figure}

We further analyze the sensitivity of MUSE to three key hyper-parameters under the forecasting horizon $H=720$. The main observations are as follows. 1) Figure~\ref{fig:parameter_sensitivity}(a) shows the effect of the look-back window length $L$. Increasing the historical context generally improves forecasting performance, but the performance is non-monotonic in $L$, and the preferred range is dataset-specific. A longer look-back window provides richer historical context while compressing a longer temporal span into the fixed visual grid, making the choice of $L$ depend on dataset-specific temporal characteristics. 2) Figure~\ref{fig:parameter_sensitivity}(b) studies the low-dimensional feature dimension $D$ in VCR. Performance varies only marginally over $D\in\{64,128,192,256\}$, while the default $D=192$ achieves the best overall performance, indicating that VCR is robust to this dimension and requires little fine-grained tuning. 3) Figure~\ref{fig:parameter_sensitivity}(c) examines the TP bottleneck dimension $D_{\mathrm{tp}}$, i.e., the latent dimension of the low-dimensional bottleneck in the Temporal--Periodic Adapter introduced in Section~\ref{sec:tpr}. Performance remains stable over $D_{\mathrm{tp}}\in\{16,32,64,128\}$, with the default $D_{\mathrm{tp}}=64$ achieving the best overall performance. This shows that TPR effectively refines temporal--periodic structures with a compact bottleneck and is robust to $D_{\mathrm{tp}}$. Overall, MUSE is robust to both adaptation dimensions, while the look-back window shows dataset-dependent sensitivity due to interactions between temporal characteristics and the fixed visual representation.

\section{Conclusions}

In this paper, we propose MUSE, a dependency-aware adaptation framework for multivariate time-series forecasting with a fully frozen pretrained MAE. Specifically, the Variable Context Refinement Module (VCR) aggregates intra-variable temporal information and models cross-variable dependencies while preserving independent visual representation spaces. The Temporal--Periodic Refinement Module (TPR) performs lightweight refinement across encoder depths and explicitly captures across-period temporal and within-period periodic dependencies. The predictions from the two modules are fused through a learnable prediction-level gate. Experiments on 10 real-world datasets demonstrate that MUSE achieves state-of-the-art performance, validating dependency-aware adaptation of frozen visual representations.

% ==================================================
% Required and recommended statements
% ==================================================

% \input{sections/06_ai_statement}
% \input{sections/07_ethics}
% \input{sections/08_reproducibility}

% ==================================================
% References
% ==================================================

\bibliography{references}
\bibliographystyle{iclr2027_conference}

% ==================================================
% Appendix
% ==================================================

\clearpage
\appendix
\section{Details of the Forward Module}
\label{app:forward_module}

MUSE follows the period-based imaging, spatial alignment, and deterministic masking procedure of VisionTS~\citep{chen2025visionts}. Let $F$ denote the periodicity, $I\times I$ the input resolution of the pretrained MAE, and $p$ the patch size, such that $G=I/p$ patches lie along each spatial axis. Following VisionTS, we set $\epsilon=10^{-5}$, $c_{\mathrm{norm}}=0.4$, and $c_{\mathrm{align}}=0.4$, and use bilinear interpolation for image resizing. The complete procedure for constructing the visible historical tokens is summarized in Algorithm~\ref{alg:forward_module}.

\begin{algorithm}[H]
\caption{Forward Module of MUSE}
\label{alg:forward_module}
\begin{algorithmic}[1]
\Statex \textbf{Input:} Historical series $\mathbf{X}\in\mathbb{R}^{L\times V}$, forecasting horizon $H$, periodicity $F$, image size $I$, patch size $p$
\Statex \textbf{Output:} Visible historical tokens $\mathbf{T}\in\mathbb{R}^{V\times P\times E}$

\State $G \gets I/p$

\Statex \textbf{Variable-wise normalization}
\For{$v=1,\ldots,V$}
    \State $\mu_v \gets \operatorname{Mean}(\mathbf{X}_{:,v})$
    \State $\sigma_v \gets
    \sqrt{\operatorname{Var}(\mathbf{X}_{:,v}-\mu_v)+\epsilon}$
    \State $\widetilde{\mathbf{x}}^{(v)}
    \gets c_{\mathrm{norm}}(\mathbf{X}_{:,v}-\mu_v)/\sigma_v$
\EndFor

\Statex \textbf{Image geometry and alignment}
\State $\ell_{\mathrm{left}}
\gets (F-(L\bmod F))\bmod F$
\State $\ell_{\mathrm{right}}
\gets (F-(H\bmod F))\bmod F$
\State $\rho \gets
\dfrac{L+\ell_{\mathrm{left}}}
{L+\ell_{\mathrm{left}}+H+\ell_{\mathrm{right}}}$
\State $K_{\mathrm{in}}
\gets \max\!\left(
1,
\left\lfloor
\rho G c_{\mathrm{align}}
\right\rfloor
\right)$
\State $K_{\mathrm{out}}\gets G-K_{\mathrm{in}}$
\State $P\gets GK_{\mathrm{in}}$

\Statex \textbf{Period-based imaging}
\For{$v=1,\ldots,V$}
    \State $\widetilde{\mathbf{x}}_{\mathrm{pad}}^{(v)}
    \gets
    \operatorname{ReplicatePadLeft}
    \left(
    \widetilde{\mathbf{x}}^{(v)},
    \ell_{\mathrm{left}}
    \right)$
    
    \State $\mathbf{A}^{(v)}
    \gets
    \operatorname{Reshape}
    \left(
    \widetilde{\mathbf{x}}_{\mathrm{pad}}^{(v)},
    F,
    \frac{L+\ell_{\mathrm{left}}}{F}
    \right)$
    
    \State $\mathbf{A}^{(v)}
    \gets
    \operatorname{BilinearResize}
    \left(
    \mathbf{A}^{(v)},
    I,
    K_{\mathrm{in}}p
    \right)$
    
    \State $\mathbf{I}_{\mathrm{gray}}^{(v)}
    \gets
    \operatorname{Concat}
    \left(
    \mathbf{A}^{(v)},
    \mathbf{0}_{I\times K_{\mathrm{out}}p}
    \right)$
    
    \State $\mathbf{I}_{\mathrm{rgb}}^{(v)}
    \gets
    \operatorname{RepeatChannels}
    \left(
    \mathbf{I}_{\mathrm{gray}}^{(v)},
    3
    \right)$
\EndFor

\Statex \textbf{Deterministic masking and visual embedding}
\State Construct $\boldsymbol{\Omega}\in\{0,1\}^{G\times G}$ such that
$\Omega_{r,c}=0$ for $c\leq K_{\mathrm{in}}$ and
$\Omega_{r,c}=1$ otherwise

\For{$v=1,\ldots,V$}
    \State $\mathbf{U}^{(v)}
    \gets
    \operatorname{FrozenPatchEmbed}
    \left(
    \mathbf{I}_{\mathrm{rgb}}^{(v)}
    \right)
    +\mathbf{E}_{\mathrm{pos}}$
    
    \State $\mathbf{T}_{v}
    \gets
    \operatorname{SelectVisible}
    \left(
    \mathbf{U}^{(v)},
    \boldsymbol{\Omega}
    \right)
    \in\mathbb{R}^{P\times E}$
\EndFor

\State \Return
$\mathbf{T}
=
[\mathbf{T}_{1},\ldots,\mathbf{T}_{V}]
\in\mathbb{R}^{V\times P\times E}$

\end{algorithmic}
\end{algorithm}

The fixed mask retains the first $K_{\mathrm{in}}$ patch columns as visible historical observations and masks the remaining $K_{\mathrm{out}}$ columns, such that only historical patch tokens are retained as visible inputs.

\section{Experimental Details}
\label{app:experimental_details}

\subsection{Datasets}
\label{app:datasets}
\begin{table}[t]
    \centering
    \caption{Statistics of multivariate datasets.}
    \label{tab:dataset_statistics}
    \small
    \begin{tabular}{l l c r r c}
        \toprule
        Dataset & Domain & Frequency & Length & Dim & Split \\
        \midrule
        ETTh1       & Electricity & 1 hour  & 14,400 & 7   & 6:2:2 \\
        ETTh2       & Electricity & 1 hour  & 14,400 & 7   & 6:2:2 \\
        ETTm1       & Electricity & 15 mins & 57,600 & 7   & 6:2:2 \\
        ETTm2       & Electricity & 15 mins & 57,600 & 7   & 6:2:2 \\
        Electricity & Electricity & 1 hour  & 26,304 & 321 & 7:1:2 \\
        Wind        & Energy      & 15 mins & 48,673 & 7   & 7:1:2 \\
        Weather     & Environment & 10 mins & 52,696 & 21  & 7:1:2 \\
        AQWan       & Environment & 1 hour  & 35,064 & 11  & 6:2:2 \\
        ZafNoo      & Nature      & 30 mins & 19,225 & 11  & 7:1:2 \\
        CzeLan      & Nature      & 30 mins & 19,934 & 11  & 7:1:2 \\
        \bottomrule
    \end{tabular}
\end{table}

We conduct experiments on 10 commonly used multivariate time-series forecasting datasets, with their statistics summarized in Table~\ref{tab:dataset_statistics}. Specifically, (1) \textbf{ETT} consists of four subsets: ETTh1 and ETTh2 contain hourly observations, while ETTm1 and ETTm2 are recorded every 15 minutes. Each subset contains seven variables related to electricity transformers. (2) \textbf{Electricity} contains hourly electricity consumption records from 321 clients. (3) \textbf{Wind} contains seven wind power variables recorded at 15-minute intervals from 2020 to 2021. (4) \textbf{Weather} contains 21 meteorological variables recorded every 10 minutes throughout 2020. (5) \textbf{AQWan} contains hourly air-quality measurements from a monitoring station over four years, with 11 variables. (6) \textbf{ZafNoo} and \textbf{CzeLan} are derived from the Sapflux data project and contain sap-flow measurements and environmental variables recorded every 30 minutes, with 11 variables each. The dataset statistics and train/validation/test splits follow TFB~\citep{qiu2024tfb}.

\subsection{Implementation Details}
\label{app:implementation_details}

1) We use the Adam optimizer and MSE as the training loss. VCR, TPR, and the prediction-level gate are supervised by separate MSE losses, which are equally weighted in the overall training objective. VCR, TPR, and the layer-wise injection parameters $s_l$ use the main learning rate, while the gate uses an independent parameter group with a learning rate ten times the main learning rate. To ensure a fair comparison, we do not apply the ``Drop Last'' trick during testing~\citep{qiu2024tfb,qiu2025tab}.

2) At the beginning of training, the VCR write-back mapping is zero-initialized, while $W_{\mathrm{up}}$ in TPR is initialized with Xavier uniform initialization and the initial layer-wise injection strength is set to $0.05$. The prediction-level gate is initialized with $g=0.5$.

\section{Experimental Results}
\label{app:experimental_results}

\subsection{Limitations and Future Work}

Despite the strong forecasting performance, MUSE still has several limitations. First, the architecture requires VCR and TPR to independently process visual representations and generate predictions, leading to higher memory consumption and inference latency. Second, although both modules are designed with explicit structural motivations, the learned dependency patterns and refinement behaviors inside the model have not yet been thoroughly characterized.

Future work could improve computational efficiency by reducing redundant computation or developing more compact mechanisms for combining variable-context and temporal--periodic modeling. It would also be valuable to further investigate how VCR and TPR modify frozen visual representations. Beyond this, extending the framework to capture more flexible multi-scale temporal and dynamic periodic structures may further improve its applicability to diverse time-series characteristics across different domains.

\subsection{Full Results}
\label{app:full_results}

Full forecasting results on the 10 multivariate time-series datasets under all forecasting horizons are reported in Table~\ref{tab:main_results_full}. The corresponding average results are summarized in Table~\ref{tab:main_results_avg} in the main text.

\begin{table}[H]
    \centering
    \caption{Full forecasting results on 10 multivariate time-series datasets under horizons $H\in\{96,192,336,720\}$. \textcolor{red}{\textbf{Red}}: the best, \textcolor{blue}{\underline{Blue}}: the 2nd best. Avg denotes the average over the four forecasting horizons. Lower is better.}
    \label{tab:main_results_full}
    \scriptsize
    \setlength{\tabcolsep}{1.15pt}
    \renewcommand{\arraystretch}{1.04}
    \resizebox{\textwidth}{!}{%
    \begin{tabular}{cc*{9}{cc}}
        \toprule
        \multicolumn{2}{c}{\textbf{Models}}
        & \multicolumn{2}{c}{MUSE (ours)}
        & \multicolumn{2}{c}{DMMV-A}
        & \multicolumn{2}{c}{VisionTS}
        & \multicolumn{2}{c}{Time-LLM}
        & \multicolumn{2}{c}{GPT4TS}
        & \multicolumn{2}{c}{DLinear}
        & \multicolumn{2}{c}{PatchTST}
        & \multicolumn{2}{c}{TimesNet}
        & \multicolumn{2}{c}{FEDformer} \\
        \multicolumn{2}{c}{\textbf{Metrics}}
        & MSE & MAE
        & MSE & MAE
        & MSE & MAE
        & MSE & MAE
        & MSE & MAE
        & MSE & MAE
        & MSE & MAE
        & MSE & MAE
        & MSE & MAE \\
        \midrule

        \multirow{5}{*}{\rotatebox[origin=c]{90}{ETTh1}} & 96 & \textcolor{red}{\textbf{0.338}} & \textcolor{red}{\textbf{0.371}} & \textcolor{blue}{\underline{0.354}} & 0.389 & 0.357 & \textcolor{blue}{\underline{0.388}} & 0.376 & 0.402 & 0.370 & 0.389 & 0.375 & 0.399 & 0.370 & 0.399 & 0.384 & 0.402 & 0.376 & 0.419 \\
        & 192 & \textcolor{blue}{\underline{0.388}} & \textcolor{red}{\textbf{0.403}} & 0.393 & \textcolor{blue}{\underline{0.405}} & \textcolor{red}{\textbf{0.387}} & 0.409 & 0.407 & 0.421 & 0.412 & 0.413 & 0.405 & 0.416 & 0.413 & 0.421 & 0.436 & 0.429 & 0.420 & 0.448 \\
        & 336 & 0.405 & 0.417 & \textcolor{red}{\textbf{0.387}} & \textcolor{red}{\textbf{0.413}} & \textcolor{blue}{\underline{0.392}} & \textcolor{blue}{\underline{0.414}} & 0.430 & 0.438 & 0.448 & 0.431 & 0.439 & 0.443 & 0.422 & 0.436 & 0.491 & 0.469 & 0.459 & 0.465 \\
        & 720 & \textcolor{red}{\textbf{0.407}} & \textcolor{red}{\textbf{0.431}} & 0.445 & 0.450 & \textcolor{blue}{\underline{0.413}} & \textcolor{blue}{\underline{0.443}} & 0.457 & 0.468 & 0.441 & 0.449 & 0.472 & 0.490 & 0.447 & 0.466 & 0.521 & 0.500 & 0.506 & 0.507 \\
        \cmidrule{2-20}
        & Avg & \textcolor{red}{\textbf{0.384}} & \textcolor{red}{\textbf{0.405}} & 0.395 & \textcolor{blue}{\underline{0.414}} & \textcolor{blue}{\underline{0.388}} & \textcolor{blue}{\underline{0.414}} & 0.418 & 0.432 & 0.418 & 0.421 & 0.423 & 0.437 & 0.413 & 0.431 & 0.458 & 0.450 & 0.440 & 0.460 \\
        \midrule

        \multirow{5}{*}{\rotatebox[origin=c]{90}{ETTh2}} & 96 & \textcolor{red}{\textbf{0.264}} & \textcolor{red}{\textbf{0.325}} & 0.294 & 0.349 & \textcolor{red}{\textbf{0.264}} & \textcolor{red}{\textbf{0.325}} & 0.286 & 0.346 & 0.280 & \textcolor{blue}{\underline{0.335}} & 0.289 & 0.353 & \textcolor{blue}{\underline{0.274}} & 0.336 & 0.340 & 0.374 & 0.358 & 0.397 \\
        & 192 & \textcolor{red}{\textbf{0.320}} & \textcolor{red}{\textbf{0.363}} & 0.339 & 0.395 & \textcolor{blue}{\underline{0.321}} & \textcolor{blue}{\underline{0.368}} & 0.361 & 0.391 & 0.348 & 0.380 & 0.383 & 0.418 & 0.339 & 0.379 & 0.402 & 0.414 & 0.429 & 0.439 \\
        & 336 & 0.342 & 0.385 & \textcolor{red}{\textbf{0.322}} & \textcolor{blue}{\underline{0.384}} & 0.386 & 0.417 & 0.390 & 0.414 & 0.380 & 0.405 & 0.448 & 0.465 & \textcolor{blue}{\underline{0.329}} & \textcolor{red}{\textbf{0.380}} & 0.452 & 0.452 & 0.496 & 0.487 \\
        & 720 & \textcolor{red}{\textbf{0.373}} & \textcolor{red}{\textbf{0.416}} & 0.392 & 0.425 & 0.387 & 0.426 & 0.405 & 0.434 & 0.406 & 0.436 & 0.605 & 0.551 & \textcolor{blue}{\underline{0.379}} & \textcolor{blue}{\underline{0.422}} & 0.462 & 0.468 & 0.463 & 0.474 \\
        \cmidrule{2-20}
        & Avg & \textcolor{red}{\textbf{0.325}} & \textcolor{red}{\textbf{0.372}} & 0.337 & 0.388 & 0.339 & 0.384 & 0.361 & 0.396 & 0.354 & 0.389 & 0.431 & 0.447 & \textcolor{blue}{\underline{0.330}} & \textcolor{blue}{\underline{0.379}} & 0.414 & 0.427 & 0.437 & 0.449 \\
        \midrule

        \multirow{5}{*}{\rotatebox[origin=c]{90}{ETTm1}} & 96 & \textcolor{blue}{\underline{0.285}} & 0.330 & \textcolor{red}{\textbf{0.279}} & \textcolor{blue}{\underline{0.329}} & 0.289 & \textcolor{red}{\textbf{0.326}} & 0.291 & 0.341 & 0.300 & 0.340 & 0.299 & 0.343 & 0.290 & 0.342 & 0.338 & 0.375 & 0.379 & 0.419 \\
        & 192 & \textcolor{red}{\textbf{0.317}} & \textcolor{red}{\textbf{0.353}} & \textcolor{red}{\textbf{0.317}} & \textcolor{blue}{\underline{0.357}} & \textcolor{blue}{\underline{0.325}} & \textcolor{red}{\textbf{0.353}} & 0.341 & 0.369 & 0.343 & 0.368 & 0.335 & 0.365 & 0.332 & 0.369 & 0.374 & 0.387 & 0.426 & 0.441 \\
        & 336 & \textcolor{red}{\textbf{0.338}} & \textcolor{red}{\textbf{0.369}} & 0.351 & 0.381 & \textcolor{blue}{\underline{0.349}} & \textcolor{blue}{\underline{0.377}} & 0.359 & 0.379 & 0.376 & 0.386 & 0.369 & 0.386 & 0.366 & 0.392 & 0.410 & 0.411 & 0.445 & 0.459 \\
        & 720 & \textcolor{red}{\textbf{0.379}} & \textcolor{red}{\textbf{0.397}} & 0.411 & 0.415 & \textcolor{blue}{\underline{0.388}} & \textcolor{blue}{\underline{0.412}} & 0.433 & 0.419 & 0.431 & 0.416 & 0.425 & 0.421 & 0.416 & 0.420 & 0.478 & 0.450 & 0.543 & 0.490 \\
        \cmidrule{2-20}
        & Avg & \textcolor{red}{\textbf{0.330}} & \textcolor{red}{\textbf{0.362}} & 0.340 & 0.371 & \textcolor{blue}{\underline{0.338}} & \textcolor{blue}{\underline{0.367}} & 0.356 & 0.377 & 0.363 & 0.378 & 0.357 & 0.379 & 0.351 & 0.381 & 0.400 & 0.406 & 0.448 & 0.452 \\
        \midrule

        \multirow{5}{*}{\rotatebox[origin=c]{90}{ETTm2}} & 96 & \textcolor{red}{\textbf{0.162}} & 0.252 & 0.172 & 0.260 & \textcolor{blue}{\underline{0.163}} & 0.251 & \textcolor{red}{\textbf{0.162}} & \textcolor{red}{\textbf{0.248}} & \textcolor{blue}{\underline{0.163}} & \textcolor{blue}{\underline{0.249}} & 0.167 & 0.269 & 0.165 & 0.255 & 0.187 & 0.267 & 0.203 & 0.287 \\
        & 192 & \textcolor{blue}{\underline{0.222}} & 0.293 & 0.227 & 0.298 & 0.233 & 0.304 & 0.235 & 0.304 & \textcolor{blue}{\underline{0.222}} & \textcolor{red}{\textbf{0.291}} & 0.224 & 0.303 & \textcolor{red}{\textbf{0.220}} & \textcolor{blue}{\underline{0.292}} & 0.249 & 0.309 & 0.269 & 0.328 \\
        & 336 & \textcolor{red}{\textbf{0.270}} & \textcolor{red}{\textbf{0.321}} & \textcolor{blue}{\underline{0.272}} & \textcolor{blue}{\underline{0.327}} & 0.304 & 0.349 & 0.280 & 0.329 & 0.273 & \textcolor{blue}{\underline{0.327}} & 0.281 & 0.342 & 0.274 & 0.329 & 0.321 & 0.351 & 0.325 & 0.366 \\
        & 720 & \textcolor{red}{\textbf{0.327}} & \textcolor{red}{\textbf{0.371}} & \textcolor{blue}{\underline{0.351}} & 0.381 & 0.361 & 0.391 & 0.366 & 0.382 & 0.357 & \textcolor{blue}{\underline{0.376}} & 0.397 & 0.421 & 0.362 & 0.385 & 0.408 & 0.403 & 0.421 & 0.415 \\
        \cmidrule{2-20}
        & Avg & \textcolor{red}{\textbf{0.245}} & \textcolor{red}{\textbf{0.309}} & 0.256 & 0.317 & 0.265 & 0.324 & 0.261 & 0.316 & \textcolor{blue}{\underline{0.254}} & \textcolor{blue}{\underline{0.311}} & 0.267 & 0.334 & 0.255 & 0.315 & 0.291 & 0.333 & 0.305 & 0.349 \\
        \midrule

        \multirow{5}{*}{\rotatebox[origin=c]{90}{Electricity}} & 96 & \textcolor{blue}{\underline{0.129}} & \textcolor{blue}{\underline{0.221}} & \textcolor{red}{\textbf{0.126}} & \textcolor{red}{\textbf{0.213}} & 0.133 & 0.225 & 0.137 & 0.233 & 0.141 & 0.239 & 0.140 & 0.237 & \textcolor{blue}{\underline{0.129}} & 0.222 & 0.168 & 0.272 & 0.193 & 0.308 \\
        & 192 & \textcolor{blue}{\underline{0.146}} & \textcolor{blue}{\underline{0.240}} & \textcolor{red}{\textbf{0.145}} & \textcolor{red}{\textbf{0.237}} & 0.151 & 0.245 & 0.152 & 0.247 & 0.158 & 0.253 & 0.153 & 0.249 & 0.157 & \textcolor{blue}{\underline{0.240}} & 0.184 & 0.289 & 0.201 & 0.315 \\
        & 336 & 0.164 & 0.260 & \textcolor{red}{\textbf{0.162}} & \textcolor{red}{\textbf{0.254}} & 0.171 & 0.266 & 0.169 & 0.267 & 0.172 & 0.266 & 0.169 & 0.267 & \textcolor{blue}{\underline{0.163}} & \textcolor{blue}{\underline{0.259}} & 0.198 & 0.300 & 0.214 & 0.329 \\
        & 720 & \textcolor{red}{\textbf{0.190}} & \textcolor{red}{\textbf{0.283}} & \textcolor{blue}{\underline{0.197}} & \textcolor{blue}{\underline{0.286}} & 0.200 & 0.293 & 0.200 & 0.290 & 0.207 & 0.293 & 0.203 & 0.301 & \textcolor{blue}{\underline{0.197}} & 0.290 & 0.220 & 0.320 & 0.246 & 0.355 \\
        \cmidrule{2-20}
        & Avg & \textcolor{red}{\textbf{0.157}} & \textcolor{blue}{\underline{0.251}} & \textcolor{blue}{\underline{0.158}} & \textcolor{red}{\textbf{0.248}} & 0.164 & 0.257 & 0.165 & 0.259 & 0.170 & 0.263 & 0.166 & 0.264 & 0.162 & 0.253 & 0.193 & 0.295 & 0.214 & 0.327 \\
        \midrule

        \multirow{5}{*}{\rotatebox[origin=c]{90}{Wind}} & 96 & \textcolor{red}{\textbf{0.864}} & 0.643 & 0.977 & 0.672 & 0.984 & 0.675 & 0.955 & 0.674 & 0.952 & \textcolor{blue}{\underline{0.638}} & \textcolor{blue}{\underline{0.881}} & \textcolor{red}{\textbf{0.632}} & 0.889 & 0.652 & 0.998 & 0.658 & 1.030 & 0.697 \\
        & 192 & \textcolor{red}{\textbf{0.998}} & \textcolor{blue}{\underline{0.727}} & 1.125 & 0.747 & 1.091 & 0.741 & 1.178 & 0.771 & 1.157 & 0.742 & \textcolor{blue}{\underline{1.034}} & \textcolor{red}{\textbf{0.715}} & 1.076 & 0.747 & 1.214 & 0.752 & 1.275 & 0.807 \\
        & 336 & \textcolor{red}{\textbf{1.074}} & \textcolor{red}{\textbf{0.769}} & 1.218 & 0.780 & 1.202 & 0.807 & 1.236 & 0.818 & 1.328 & 0.819 & \textcolor{blue}{\underline{1.159}} & \textcolor{blue}{\underline{0.779}} & 1.209 & 0.809 & 1.318 & 0.828 & 1.307 & 0.864 \\
        & 720 & \textcolor{red}{\textbf{1.135}} & \textcolor{red}{\textbf{0.800}} & 1.240 & \textcolor{blue}{\underline{0.805}} & \textcolor{blue}{\underline{1.207}} & 0.817 & 1.322 & 0.855 & 1.441 & 0.873 & 1.233 & 0.815 & 1.304 & 0.851 & 1.474 & 0.887 & 1.402 & 0.890 \\
        \cmidrule{2-20}
        & Avg & \textcolor{red}{\textbf{1.018}} & \textcolor{red}{\textbf{0.735}} & 1.140 & \textcolor{blue}{\underline{0.751}} & 1.121 & 0.760 & 1.173 & 0.780 & 1.220 & 0.768 & \textcolor{blue}{\underline{1.077}} & \textcolor{red}{\textbf{0.735}} & 1.120 & 0.765 & 1.251 & 0.781 & 1.254 & 0.815 \\
        \midrule

        \multirow{5}{*}{\rotatebox[origin=c]{90}{Weather}} & 96 & \textcolor{red}{\textbf{0.142}} & 0.197 & \textcolor{blue}{\underline{0.143}} & \textcolor{blue}{\underline{0.195}} & 0.156 & 0.210 & 0.155 & 0.199 & 0.148 & \textcolor{red}{\textbf{0.188}} & 0.176 & 0.237 & 0.149 & 0.198 & 0.172 & 0.220 & 0.217 & 0.296 \\
        & 192 & \textcolor{blue}{\underline{0.188}} & 0.242 & \textcolor{red}{\textbf{0.187}} & 0.242 & 0.229 & 0.275 & 0.223 & 0.261 & 0.192 & \textcolor{red}{\textbf{0.230}} & 0.220 & 0.282 & 0.194 & \textcolor{blue}{\underline{0.241}} & 0.219 & 0.261 & 0.276 & 0.336 \\
        & 336 & \textcolor{red}{\textbf{0.233}} & \textcolor{blue}{\underline{0.279}} & \textcolor{blue}{\underline{0.237}} & \textcolor{red}{\textbf{0.273}} & 0.297 & 0.319 & 0.251 & \textcolor{blue}{\underline{0.279}} & 0.246 & \textcolor{red}{\textbf{0.273}} & 0.265 & 0.319 & 0.245 & 0.282 & 0.280 & 0.306 & 0.339 & 0.380 \\
        & 720 & \textcolor{red}{\textbf{0.299}} & \textcolor{blue}{\underline{0.326}} & \textcolor{blue}{\underline{0.302}} & \textcolor{red}{\textbf{0.315}} & 0.366 & 0.368 & 0.345 & 0.342 & 0.320 & 0.328 & 0.333 & 0.362 & 0.314 & 0.334 & 0.365 & 0.359 & 0.403 & 0.428 \\
        \cmidrule{2-20}
        & Avg & \textcolor{red}{\textbf{0.216}} & 0.261 & \textcolor{blue}{\underline{0.217}} & \textcolor{blue}{\underline{0.256}} & 0.262 & 0.293 & 0.244 & 0.270 & 0.227 & \textcolor{red}{\textbf{0.255}} & 0.249 & 0.300 & 0.226 & 0.264 & 0.259 & 0.287 & 0.309 & 0.360 \\
        \midrule

        \multirow{5}{*}{\rotatebox[origin=c]{90}{AQWan}} & 96 & \textcolor{blue}{\underline{0.739}} & \textcolor{blue}{\underline{0.465}} & 0.762 & 0.506 & 0.760 & 0.481 & \textcolor{red}{\textbf{0.712}} & 0.468 & 0.745 & \textcolor{red}{\textbf{0.464}} & 0.756 & 0.481 & 0.745 & 0.470 & 0.791 & 0.488 & 0.796 & 0.508 \\
        & 192 & 0.791 & \textcolor{red}{\textbf{0.487}} & \textcolor{blue}{\underline{0.789}} & 0.520 & 0.858 & 0.516 & 0.838 & 0.533 & 0.813 & \textcolor{blue}{\underline{0.490}} & 0.800 & 0.502 & 0.792 & 0.491 & \textcolor{red}{\textbf{0.779}} & \textcolor{blue}{\underline{0.490}} & 0.825 & 0.517 \\
        & 336 & 0.809 & \textcolor{red}{\textbf{0.497}} & \textcolor{blue}{\underline{0.805}} & 0.529 & 0.820 & 0.507 & \textcolor{red}{\textbf{0.788}} & 0.504 & 0.818 & \textcolor{red}{\textbf{0.497}} & 0.823 & 0.516 & 0.819 & \textcolor{blue}{\underline{0.503}} & 0.814 & 0.505 & 0.863 & 0.537 \\
        & 720 & 0.883 & 0.526 & \textcolor{red}{\textbf{0.859}} & 0.557 & 0.865 & \textcolor{blue}{\underline{0.524}} & \textcolor{blue}{\underline{0.863}} & 0.536 & 0.898 & 0.526 & 0.891 & 0.548 & 0.890 & 0.533 & 0.869 & \textcolor{red}{\textbf{0.519}} & 0.907 & 0.552 \\
        \cmidrule{2-20}
        & Avg & 0.805 & \textcolor{red}{\textbf{0.494}} & \textcolor{blue}{\underline{0.804}} & 0.528 & 0.826 & 0.507 & \textcolor{red}{\textbf{0.800}} & 0.510 & 0.819 & \textcolor{red}{\textbf{0.494}} & 0.818 & 0.512 & 0.812 & \textcolor{blue}{\underline{0.499}} & 0.813 & 0.501 & 0.848 & 0.529 \\
        \midrule

        \multirow{5}{*}{\rotatebox[origin=c]{90}{ZafNoo}} & 96 & \textcolor{red}{\textbf{0.423}} & \textcolor{red}{\textbf{0.392}} & 0.436 & 0.410 & 0.465 & 0.411 & 0.470 & 0.436 & 0.452 & \textcolor{blue}{\underline{0.399}} & \textcolor{blue}{\underline{0.434}} & 0.411 & 0.444 & 0.426 & 0.479 & 0.424 & 0.475 & 0.441 \\
        & 192 & \textcolor{red}{\textbf{0.475}} & \textcolor{red}{\textbf{0.422}} & 0.488 & \textcolor{blue}{\underline{0.440}} & 0.537 & 0.451 & 0.610 & 0.502 & 0.526 & 0.444 & \textcolor{blue}{\underline{0.484}} & 0.444 & 0.498 & 0.456 & 0.491 & 0.446 & 0.544 & 0.479 \\
        & 336 & \textcolor{red}{\textbf{0.513}} & \textcolor{red}{\textbf{0.446}} & 0.537 & 0.466 & 0.574 & 0.470 & 0.556 & 0.482 & 0.576 & 0.472 & \textcolor{blue}{\underline{0.518}} & \textcolor{blue}{\underline{0.464}} & 0.530 & 0.480 & 0.551 & 0.479 & 0.595 & 0.517 \\
        & 720 & \textcolor{blue}{\underline{0.564}} & \textcolor{red}{\textbf{0.468}} & 0.566 & 0.483 & 0.575 & \textcolor{blue}{\underline{0.477}} & 0.608 & 0.512 & 0.634 & 0.507 & \textcolor{red}{\textbf{0.548}} & 0.486 & 0.574 & 0.499 & 0.627 & 0.511 & 0.697 & 0.560 \\
        \cmidrule{2-20}
        & Avg & \textcolor{red}{\textbf{0.494}} & \textcolor{red}{\textbf{0.432}} & 0.507 & \textcolor{blue}{\underline{0.450}} & 0.538 & 0.452 & 0.561 & 0.483 & 0.547 & 0.456 & \textcolor{blue}{\underline{0.496}} & 0.451 & 0.512 & 0.465 & 0.537 & 0.465 & 0.578 & 0.499 \\
        \midrule

        \multirow{5}{*}{\rotatebox[origin=c]{90}{CzeLan}} & 96 & \textcolor{red}{\textbf{0.170}} & \textcolor{red}{\textbf{0.224}} & 0.184 & 0.255 & 0.180 & 0.242 & 0.191 & 0.249 & 0.195 & \textcolor{blue}{\underline{0.233}} & 0.211 & 0.289 & 0.183 & 0.251 & \textcolor{blue}{\underline{0.176}} & 0.237 & 0.231 & 0.310 \\
        & 192 & \textcolor{red}{\textbf{0.203}} & \textcolor{red}{\textbf{0.255}} & \textcolor{blue}{\underline{0.207}} & 0.272 & 0.237 & 0.281 & 0.213 & 0.269 & 0.233 & \textcolor{blue}{\underline{0.259}} & 0.252 & 0.323 & 0.208 & 0.271 & 0.215 & 0.279 & 0.268 & 0.339 \\
        & 336 & \textcolor{blue}{\underline{0.234}} & \textcolor{red}{\textbf{0.283}} & 0.236 & 0.304 & 0.265 & 0.308 & 0.239 & 0.299 & 0.264 & \textcolor{red}{\textbf{0.283}} & 0.317 & 0.366 & 0.243 & 0.302 & \textcolor{red}{\textbf{0.224}} & \textcolor{blue}{\underline{0.288}} & 0.298 & 0.361 \\
        & 720 & \textcolor{red}{\textbf{0.238}} & \textcolor{red}{\textbf{0.295}} & 0.275 & 0.345 & 0.284 & 0.330 & \textcolor{blue}{\underline{0.272}} & 0.321 & 0.300 & \textcolor{blue}{\underline{0.311}} & 0.358 & 0.392 & 0.273 & 0.335 & 0.282 & 0.337 & 0.416 & 0.446 \\
        \cmidrule{2-20}
        & Avg & \textcolor{red}{\textbf{0.211}} & \textcolor{red}{\textbf{0.264}} & 0.226 & 0.294 & 0.241 & 0.290 & 0.229 & 0.285 & 0.248 & \textcolor{blue}{\underline{0.271}} & 0.285 & 0.343 & 0.227 & 0.290 & \textcolor{blue}{\underline{0.224}} & 0.285 & 0.303 & 0.364 \\
        \midrule

        \multicolumn{2}{c}{1$^{\mathrm{st}}$ Count} & \textcolor{red}{\textbf{34}} & \textcolor{red}{\textbf{32}} & \textcolor{blue}{\underline{9}} & 7 & 2 & 3 & 4 & 1 & 0 & \textcolor{blue}{\underline{9}} & 1 & 3 & 1 & 1 & 2 & 1 & 0 & 0 \\
        \bottomrule
    \end{tabular}%
    }
\end{table}

\section{Related Works}
\label{app:related_works}

\subsection{Multivariate Time Series Forecasting}

Time series are sequences of numerical observations arranged in temporal order, providing a natural representation of how real-world systems evolve over time~\citep{wu2026timeart}. They are widely observed in domains such as energy, transportation, environmental monitoring, industrial systems, and healthcare~\citep{wu2025k2vae,qiu2026dag,li2026gcgnet,qiu2026bridging,liu2026rethinking,liu2026astgi,wu2025srsnet,qiu2025dbloss,cheng2026metagnsdformer}. Time-series research spans a broad range of tasks, including forecasting~\citep{yu2025ginar,yu2025merlin}, anomaly detection~\citep{qiu2025tab,wu2025catch}, and other time-series analysis tasks. Among these, multivariate time series forecasting, which aims to predict future observations of multiple variables from their historical measurements, has been extensively studied. Early approaches mainly relied on statistical modeling~\citep{sims1980macroeconomics}, followed by machine learning methods for capturing nonlinear temporal patterns~\citep{sapankevych2009timeseries,ahmed2010empirical}. The rapid progress of deep learning in other fields, including computer vision and video generation~\citep{ma2024followpose,ma2024followyouremoji,ma2025controllable,ma2026livelight}, has demonstrated the strong representation learning capability of deep neural networks and motivated the adoption of deep learning in time series forecasting. In recent years, deep learning has become the dominant paradigm. Representative methods include LSTNet~\citep{lai2018modeling}, Autoformer~\citep{wu2021autoformer}, FEDformer~\citep{zhou2022fedformer}, DLinear~\citep{zeng2023dlinear}, PatchTST~\citep{nie2023patchtst}, TimesNet~\citep{wu2023timesnet}, iTransformer~\citep{liu2024itransformer}, Pathformer~\citep{chen2024pathformer}, TimeMixer~\citep{wang2024timemixer}, and DUET~\citep{qiu2025duet}. These methods have continuously improved forecasting performance, but are typically trained end-to-end from scratch on numerical time series, with limited use of the rich prior knowledge provided by large-scale cross-domain pretrained models.

\subsection{LVM-Based Time Series Forecasting}

With the development of visual models such as ResNet~\citep{he2016resnet}, VGG~\citep{simonyan2015vgg}, Inception~\citep{szegedy2015inception}, ViT~\citep{dosovitskiy2021vit}, Swin Transformer~\citep{liu2021swin}, and MAE~\citep{he2022mae}, transferring visual representations to time-series forecasting has attracted increasing attention. One line of work adopts channel-independent modeling. ViTime~\citep{yang2025vitime} explores a vision-based foundation model for time-series forecasting. VisionTS~\citep{chen2025visionts} reformulates forecasting as masked image reconstruction and leverages a pretrained MAE for prediction. DMMV~\citep{shen2025dmmv} builds upon VisionTS by introducing a numerical branch for complementary forecasting. Aurora~\citep{wu2026aurora} develops a cross-domain multimodal time-series foundation model for zero-shot forecasting. Another line of work adopts channel-dependent modeling. Time-VLM~\citep{zhong2025time} and VisionTS++~\citep{shen2025visiontspp} both encode multiple variables into a single image to capture inter-variable relationships.

Overall, channel-independent methods preserve a complete visual representation space for each variable but lack explicit cross-variable interaction, while channel-dependent methods capture variable relationships at the cost of sharing a limited image space across variables. MUSE addresses both limitations by constructing an individual image for each variable while explicitly modeling inter-variable dependencies, and further refines temporal and periodic structures through lightweight adaptation over a fully frozen visual backbone.

\end{document}